%% file: main.tex
\documentclass[runningheads]{llncs}

\usepackage[mobile]{eccv}

\newcommand{\framework}{\textbf{StructSplat}}

\usepackage{eccvabbrv}

\usepackage{graphicx}
\usepackage{booktabs}
\usepackage{marvosym}

\usepackage[accsupp]{axessibility}  

\usepackage{floatrow}
\newfloatcommand{capbtabbox}{table}[][\FBwidth]
\usepackage{wrapfig}

\usepackage{multirow}
\usepackage{makecell}

\usepackage{overpic}
\usepackage{caption}

\usepackage{threeparttable}

\usepackage{ulem}

\usepackage{hyperref}

\usepackage{orcidlink}

\begin{document}



\title{StructSplat: Generalizable 3D Gaussian Splatting from Uncalibrated Sparse Views} 



\titlerunning{StructSplat}

\author{Jia-Chen Zhao\inst{1,2} \and
Beiqi Chen\inst{1} \and
Xinyang Chen\inst{1}\textsuperscript{\Letter} \and Guangcong Wang\inst{2,3}\textsuperscript{\Letter} \and Liqiang Nie\inst{1}}

\authorrunning{J.-C. Zhao et al.}

\institute{Harbin Institute of Technology (Shenzhen) \and
Great Bay University \and Guangzhou CloudButterfly Technology Co., Ltd.\\
\email{jc-zhao@outlook.com}, \email{\{chenxinyang95, wanggc3, nieliqiang\}@gmail.com} \\
\begin{tabular}{l}
Project page: \textcolor{magenta}{\url{https://structsplat.github.io}} \\
Code available in: \textcolor{magenta}{\url{https://github.com/J-C-Zhao/StructSplat}}
\end{tabular}
}

\maketitle
\begingroup
\renewcommand\thefootnote{}
\footnotetext{\textsuperscript{\Letter} Corresponding authors.}
\endgroup

\begin{figure}
  \vspace{-9mm} 
    \centering
  \includegraphics[width=0.8\linewidth]{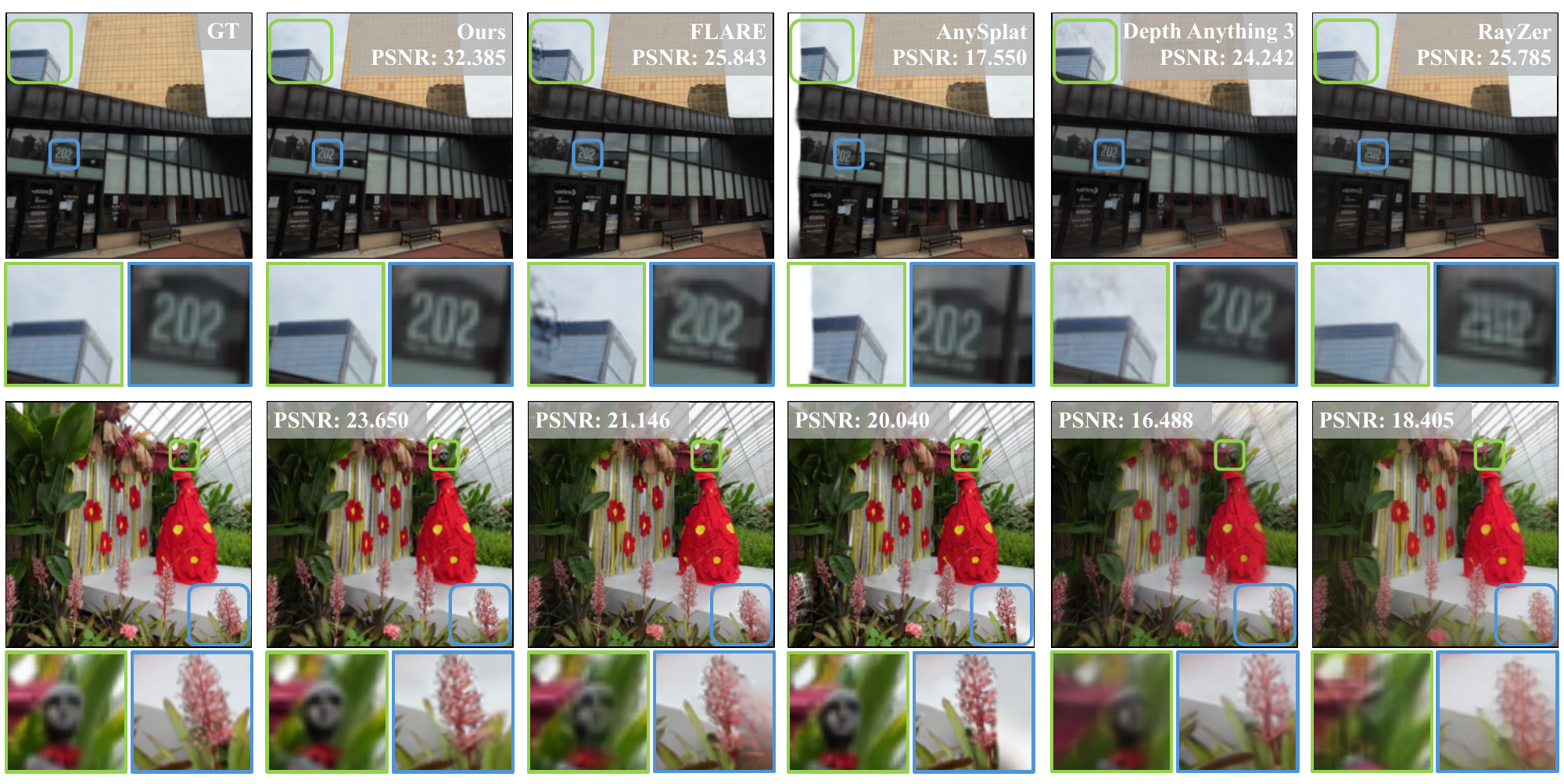}
  \caption{\textbf{Qualitative comparison for novel-view synthesis from uncalibrated sparse views.} Our method demonstrates fewer artifacts, less blur, and less shifting.
  %
  }
  \label{fig:teaser}
\vspace{-3mm} 
\end{figure}
\vspace{-10mm} 


\input{sections/00_abstract}

\input{sections/01_introction}

\input{sections/02_related}

\input{sections/03_method}

\input{sections/04_experiment}

\input{sections/05_conclusion}

\input{sections/06_acknowledgements}





%
%
\bibliographystyle{splncs04}
\bibliography{main}

\input{sections/supplementary}

\end{document}

%% file: sections/00_abstract.tex
\begin{abstract}
We present \framework, a feed-forward and generalizable 3D Gaussian reconstruction framework that operates directly on uncalibrated images without requiring camera parameters. Existing methods either rely on per-scene optimization or assume known camera poses, and often entangle geometry and appearance within a unified backbone, limiting reconstruction fidelity and generalization. Our key idea is to adopt a \textbf{structured representation} that organizes geometry, semantic, and texture cues with explicit roles in the reconstruction process. Specifically, we introduce a pixel-aligned feature injection mechanism to enable accurate texture modeling from 2D observations, incorporate semantic-aware priors to improve global consistency, and design a camera alignment strategy to prevent information leakage and improve generalization. Experiments show that our method significantly outperforms prior approaches on challenging benchmarks. 
On DL3DV, our method achieves 28.045 PSNR, surpassing AnySplat (22.377) by +5.67 dB. 
In cross-dataset evaluation, our method achieves +1.94 dB over AnySplat on ACID and +1.72 dB on RealEstate10K.

\keywords{Generalizable; 3D Gaussian Splatting; Feed-Forward}
\end{abstract}

%% file: sections/01_introction.tex
\section{Introduction}

3D Gaussian Splatting (3DGS)~\cite{kerbl20233d,fang2024mini,yu2024mip} and Neural Radiance Fields (NeRF)~\cite{mildenhall2021nerf,barron2021mip,barron2023zip,wang2023sparsenerf} have significantly advanced 3D representations and enabled high-quality novel view synthesis. To relax the dependence on accurate camera intrinsics and extrinsics, some studies~\cite{frozenrecon, noponerf, cf3dgs, inerf, nerfmm, cfnerf} explore pose-free or intrinsic-free reconstruction settings. However, they rely on \textit{per-scene optimization}, requiring iterative fitting for each individual scene. This paradigm is computationally expensive and time-consuming, and more importantly, it lacks generalization ability, as the learned representation cannot be directly transferred to unseen scenes.

Recent efforts~\cite{mvsgaussian, pixelsplat, depthsplat, gs-lrm, lvsm, mvsplat, lrm, longlrm} have moved beyond per-scene optimization toward feed-forward, generalizable novel view synthesis. One line of work~\cite{mvsgaussian, mvsplat, depthsplat, mvsnerf, geonerf} constructs 3D cost volumes via plane-sweep or homography-based warping, projecting source-view features across depth candidates for aggregation and reconstruction. Another line~\cite{pixelsplat, mvdiff, h3r, epipolartransformer, suhail2022generalizable} leverages epipolar constraints and attention to implicitly encode 3D geometry without explicit cost volumes. More recently, large-model-based methods~\cite{lrm, gs-lrm, lvsm, longlrm, lgm, instant3d} learn strong priors from large-scale data to reconstruct 3D scenes with minimal geometric constraints. Despite promising generalization, these approaches still \textit{require accurate camera intrinsics and extrinsics} obtained via SfM pipelines~\cite{colmap_sfm, glomap, dfnet, vggsfm}, limiting their applicability in real-world settings.

To address these problems, some generalizable models focus on pose-free or intrinsic-free novel view synthesis~\cite{noposplat,splatt3r,flare,da3}. For example, NoPoSplat~\cite{noposplat} addresses scale ambiguity via camera intrinsic token embedding but does not use poses. Splatt3R~\cite{splatt3r} leverages MASt3R’s~\cite{leroy2024grounding} pretrained ViT~\cite{vit} encoder and cross-attention transformers to encode uncalibrated images and predict Gaussian parameters. FLARE~\cite{flare} adopts a cascaded paradigm where camera poses are first estimated and then used to guide geometry and appearance learning. Depth Anything 3~\cite{da3} performs feed-forward reconstruction by introducing a DPT head after the geometry backbone for pixel-aligned Gaussian prediction with a pose-adaptive design.

Despite recent progress, feed-forward reconstruction from uncalibrated images remains challenging. 
\textit{First}, most existing methods lack semantic-aware modeling, limiting their ability to leverage high-level priors for robust generalization. 
\textit{Second}, geometry and appearance are often entangled within a unified backbone, which is suboptimal for recovering high-frequency textures and faithful colors, as compact 3D features are inherently biased toward spatial smoothness and geometric consistency, making them insufficient for modeling fine-grained, view-dependent appearance details. 
\textit{Third}, some approaches jointly estimate camera parameters using both source and target views, and then perform feed-forward rendering in a shared coordinate system, which may introduce target-view information through cross-view attention during rendering, leading to information leakage and degraded generalization.

In this paper, we propose \framework, a feed-forward and generalizable 3D Gaussian reconstruction framework that operates directly on uncalibrated images without requiring camera intrinsics or extrinsics. 
Our approach adopts a \textbf{structured representation} that organizes geometry, semantic, and texture information with explicit roles in the reconstruction process. 
Specifically, we introduce a geometry encoder to provide reliable structural guidance, a semantic encoder to incorporate high-level priors for global consistency, and a lightweight texture encoder to capture fine-grained appearance details. 
The texture encoder further leverages a pixel-aligned feature injection mechanism, establishing a direct pathway from 2D observations to Gaussian attributes, enabling accurate recovery of high-frequency textures and consistent colors. 
Furthermore, we design a \textbf{camera alignment strategy} that estimates camera parameters with and without target views and aligns them into a unified coordinate system, enforcing a clean separation between source and target information during both training and inference, thereby preventing information leakage and improving generalization.

Overall, our main contributions are summarized as follows:
\begin{itemize}
    \item We propose \framework, a feed-forward and generalizable 3D Gaussian reconstruction framework that operates on uncalibrated images without requiring camera parameters.

    \item We introduce a structured representation that organizes geometry, semantic, and texture cues with explicit functional roles for improved reconstruction quality.

    \item We design a camera alignment strategy to prevent information leakage and ensure robust generalization under uncalibrated settings.
    
    \item Extensive experiments demonstrate that our method significantly outperforms prior approaches on challenging benchmarks.
\end{itemize}

%% file: sections/02_related.tex
\section{Related Work}
\noindent\textbf{Generalizable Novel View Synthesis.} 
To overcome the costly per-scene optimization required by NeRF~\cite{mildenhall2021nerf,barron2021mip,barron2023zip,wang2023sparsenerf} and 3DGS~\cite{kerbl20233d,fang2024mini,yu2024mip}, recent studies have shifted toward feed-forward, generalizable models~\cite{mvsgaussian, pixelsplat, depthsplat, gs-lrm, lvsm, mvsplat, lrm, longlrm}. By learning priors from large-scale datasets, these methods enable efficient inference and can be broadly categorized by their multi-view aggregation strategies.
\textit{Cost volume-based methods}~\cite{mvsgaussian, mvsplat, mvsnerf, geonerf} construct 3D cost volumes via plane-sweep or homography warping, projecting source features across discretized depths for aggregation.  DepthSplat~\cite{depthsplat} further incorporates monocular priors to improve robustness in textureless regions.
\textit{Geometry-aware transformer methods}~\cite{pixelsplat, mvdiff, h3r, epipolartransformer, suhail2022generalizable} implicitly model 3D relationships via epipolar-constrained attention, restricting interactions along 1D epipolar lines instead of constructing explicit cost volumes.
\textit{Large-model-based methods}~\cite{lrm, gs-lrm, lvsm, longlrm, lgm} learn strong priors directly from large-scale data, reducing reliance on explicit geometric modeling. For instance, GS-LRM~\cite{gs-lrm} extends LRM~\cite{lrm} to generate 3D Gaussians, while LVSM~\cite{lvsm} performs novel view synthesis from sparse inputs using transformer architectures.
Despite their efficiency, most methods still rely on pre-computed camera poses from Structure-from-Motion (SfM) pipelines~\cite{colmap_sfm, glomap, dfnet, vggsfm}, which are computationally expensive and brittle in texture-poor or low-overlap scenarios~\cite{cf3dgs}. This reliance limits the end-to-end efficiency of feed-forward models and restricts their applicability in unconstrained real-world settings.

\noindent\textbf{Pose-free or Intrinsic-free Novel View Synthesis.} 
To simplify the 3D reconstruction pipeline and remove the dependence on pre-computed camera parameters, recent studies jointly estimate scene geometry and camera poses.
\textit{Per-scene pose-free optimization methods}~\cite{frozenrecon, noponerf, cf3dgs, inerf, nerfmm, cfnerf} optimize camera parameters together with the scene representation. For example, FrozenRecon~\cite{frozenrecon} aligns monocular depth using a trainable module, while NoPe-NeRF~\cite{noponerf} leverages undistorted depth priors to constrain relative poses. CF-3DGS~\cite{cf3dgs} further adopts an incremental strategy to estimate poses and grow a global Gaussian representation. However, these methods rely on costly per-scene optimization, limiting scalability and practical deployment.
\textit{Generalizable feed-forward pose-free methods}~\cite{coponerf, selfsplat, pf3plat, pflrm, noposplat, flare, sun2026uni3r, sheng2025spatialsplat} predict 3D representations from unposed inputs in a single forward pass. SelfSplat~\cite{selfsplat} integrates self-supervised depth and pose estimation into 3DGS, while PF3plat~\cite{pf3plat} adopts a two-stage design that first predicts coarse geometry and poses before refinement. SpatialSplat~\cite{sheng2025spatialsplat} and Uni3R~\cite{sun2026uni3r} adapt feed-forward 3D frameworks for Noval-View Segmentation. Despite improved efficiency, these methods typically assume known camera intrinsics, preventing a fully end-to-end 2D-to-3D mapping.
\textit{Fully intrinsic-free and pose-free methods}~\cite{splatt3r, anysplat, rayzer, da3} further remove both intrinsic and extrinsic dependencies. For instance, AnySplat~\cite{anysplat} employs a VGGT backbone~\cite{vggt} to directly predict camera parameters and Gaussian primitives from unconstrained views, while RayZer~\cite{rayzer} encodes camera information into latent representations. Depth Anything 3~\cite{da3} introduces a 3D foundation model with a Gaussian decoding head. 
Despite these advances, existing feed-forward approaches largely under-explore rich 2D semantics and high-frequency texture details, which are crucial for improving appearance fidelity and reconstruction quality.

%% file: sections/03_method.tex
\section{Method}

\begin{figure*}[t]
  \centering
  \includegraphics[width=\linewidth]{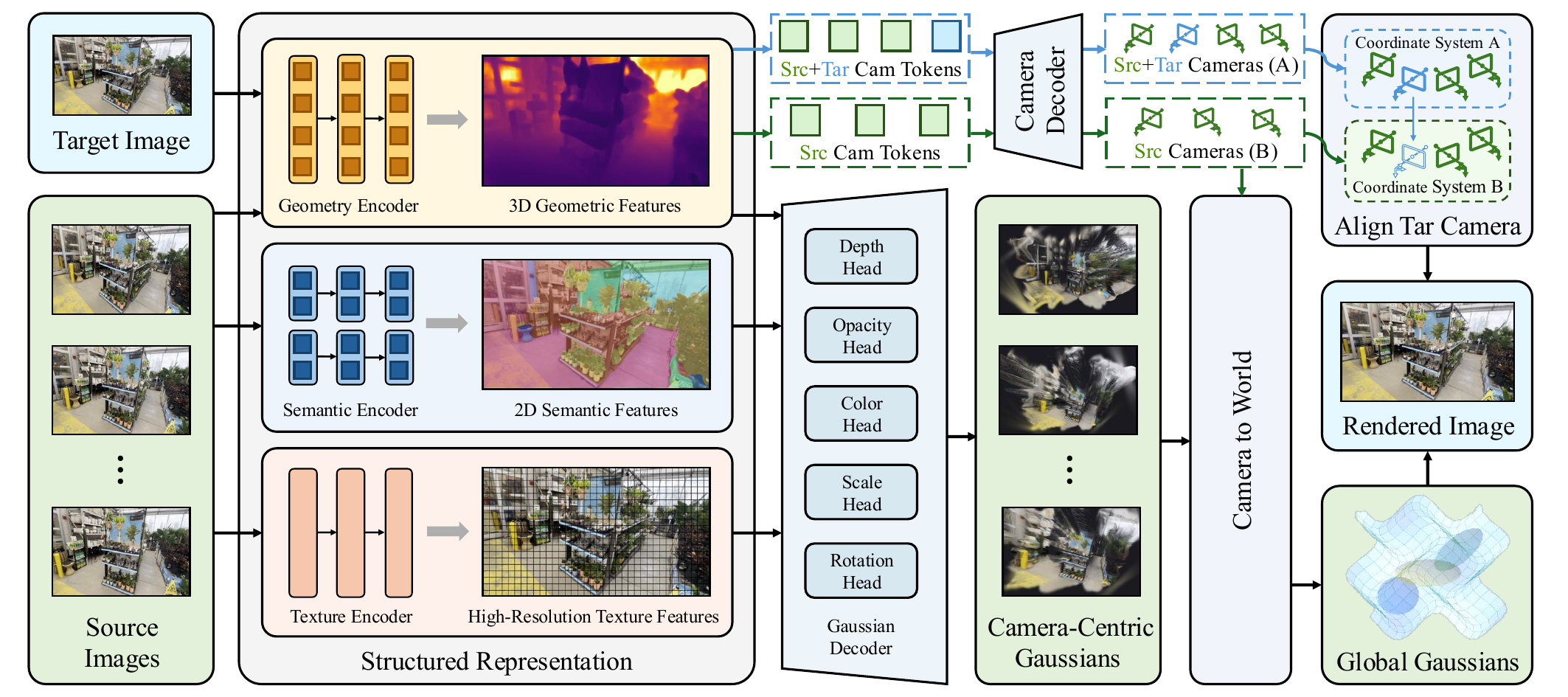}
  \caption{\textbf{Overview of our proposed StructSplat.} 
  Given uncalibrated source images, we perform feed-forward 3D reconstruction without camera parameters by adopting a structured representation that organizes texture, semantic, and geometric cues. Encoders extract multi-level features, which drive two decoding pathways: a Gaussian decoder predicts camera-centric Gaussians, while a camera decoder estimates camera parameters followed by a camera alignment module for unified registration. The aligned Gaussians are then transformed to a global representation and rendered via differentiable splatting.}
  \label{fig:overivew}
\end{figure*}

\begin{figure*}[t]
  \centering
  \includegraphics[width=\linewidth]{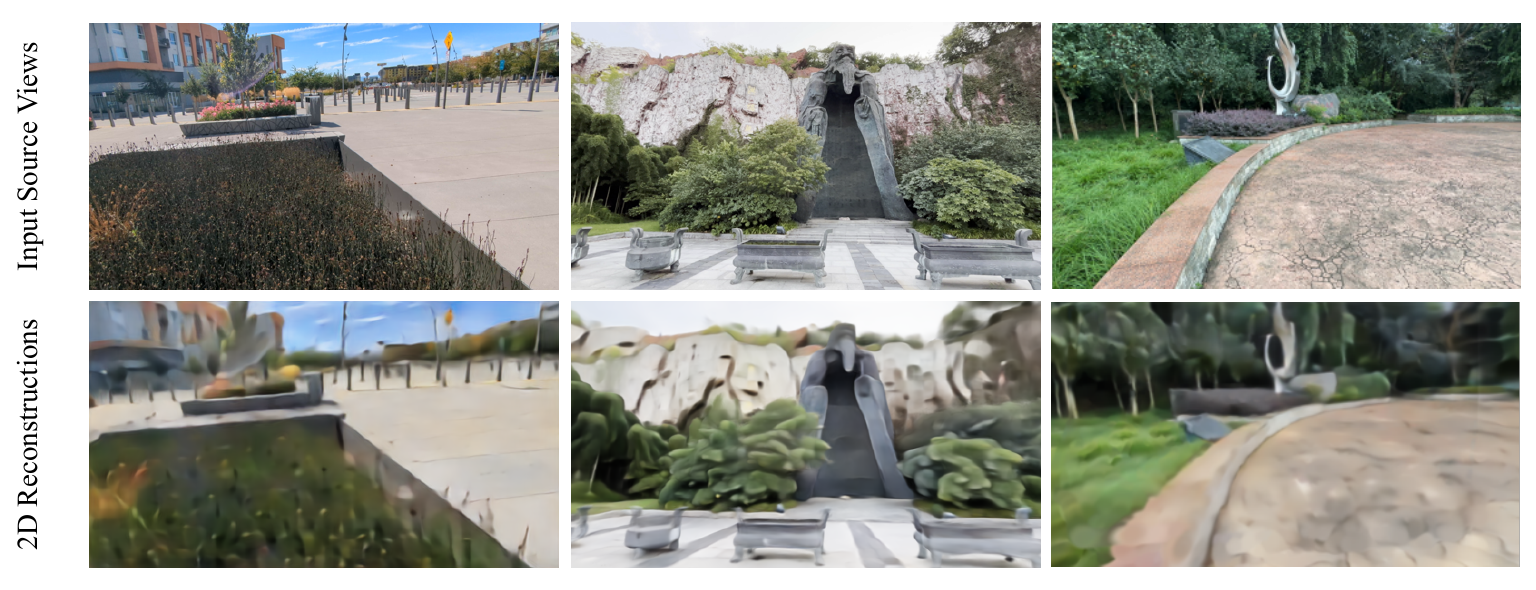}
  \caption{\textbf{Geometry-oriented features fail to capture appearance details.} We attach a color head to VGGT~\cite{vggt}, a strong pre-trained geometry encoder, to directly perform 2D reconstruction, but observe poor color fidelity and texture details even on training views. This reveals that geometry-oriented features fail to capture appearance information, motivating our structured representation with a dedicated appearance pathway.
  }
  \label{fig:motivation}
\end{figure*}

\subsection{Problem Formulation}
Given a set of uncalibrated source images $\mathcal{I}^\text{s} = \{I^\text{s}_i \in \mathbb{R}^{H \times W \times 3}\}_{i=1}^N$, our goal is to infer a 3D scene representation in the form of a set of Gaussians $\mathcal{G} = \{(\mu_j, \alpha_j, c_j, s_j, r_j)\}_{j=1}^{NMHW}$ in a feed-forward manner, which is given by $\mathcal{G} = f_\theta(\mathcal{I}^\text{s})$, where $f_\theta$ is a neural network parameterized by $\theta$. Notably, our formulation does not require camera intrinsics or extrinsics.
Each Gaussian is parameterized by its center $\mu_j$, opacity $\alpha_j$, color $c_j$, scale $s_j$, and rotation $r_j$, and we predict Gaussians per source pixel. 
Given the reconstructed Gaussians $\mathcal{G}$, novel views are synthesized via differentiable splatting and rasterization~\cite{kerbl20233d}, which is given by
$\widehat{\mathcal{I}}^\text{t} = \text{Render}(\mathcal{G}, \mathbf{p}^\text{t})$,
where $\widehat{\mathcal{I}}^\text{t}$ denotes the rendered image at target viewpoint $\mathbf{p}^\text{t}$.

\subsection{Overview of \framework}
The overview of StructSplat is illustrated in Fig. \ref{fig:overivew}. 
Given a set of uncalibrated source images, our goal is to perform feed-forward 3D reconstruction without relying on camera parameters. 
To this end, we adopt a \textbf{structured representation} that organizes multi-level visual cues for effective geometry and appearance modeling.
Specifically, the input images are processed by a set of encoders to extract complementary features, including high-resolution texture features, 2D semantic features, and 3D geometric features. 
These features are then used to drive two decoding pathways. 
The Gaussian decoder, equipped with multiple specialized heads, densely predicts the attributes of camera-centric Gaussians. 
Meanwhile, a camera decoder estimates camera parameters to resolve spatial inconsistencies across views, followed by a camera alignment module that registers all views into a unified coordinate system.
Finally, guided by the predicted cameras, a camera-to-world transformation converts camera-centric Gaussians into a global representation, which is rendered via differentiable splatting to synthesize novel views.

\subsection{Structured Generalizable 3D Representation}
\label{sec:hybrid_rep}
A straightforward way to build a generalizable 3DGS model is to attach a Gaussian head to a strong 3D geometry backbone (e.g., MASt3R~\cite{leroy2024grounding} or VGGT~\cite{vggt}). 
For example, Splatt3R~\cite{splatt3r} leverages MASt3R to encode uncalibrated images and predict Gaussian parameters, while Depth Anything 3~\cite{da3} introduces a foundation geometry model with an additional fine-tuned DPT head~\cite{dpt} to infer pixel-aligned 3D Gaussians. However, as illustrated in Fig.~\ref{fig:motivation}, directly relying on geometry features is insufficient for appearance modeling. 
When attaching a learnable color head to a strong pre-trained geometry encoder (e.g., VGGT) for 2D image reconstruction, we observe poor color fidelity and texture details, even on training views. 
This indicates a mismatch between geometry-oriented features and appearance reconstruction, since compact 3D representations are inherently biased toward spatial smoothness and geometric consistency, limiting their ability to capture fine-grained, view-dependent appearance details.

Motivated by this observation, we propose a structured representation that organizes multi-level visual cues. 
By integrating high-resolution 2D texture features and semantic priors with 3D geometry, our design preserves fine-grained appearance details while maintaining reliable structural consistency.

\noindent\textbf{Geometry Encoder.}
We employ a VGGT-based encoder to capture global geometric structure and provide reliable 3D priors for reconstruction. 
Its attention-based architecture enables effective reasoning over multi-view relationships and facilitates the aggregation of long-range contextual information across views. 
The encoder produces tokenized 3D features that encode scene geometry in a view-consistent manner, which serve as the foundation for subsequent depth and camera estimation. 
This design ensures robust geometric understanding and provides stable structural guidance for Gaussian reconstruction. We decode the 3D token features into a pixel-aligned depth map, which is then transformed into the world coordinate system to present the positions $\mu$ of the 3D Gaussians.

\noindent\textbf{Semantic Encoder.}
To complement the geometry encoder, we introduce a semantic encoder to provide high-level contextual priors for reconstruction. 
Unlike geometry features that focus on structure, semantic representations capture object-level organization and global scene understanding, which are essential for resolving ambiguities under sparse or uncalibrated views. 
Specifically, we extract semantic features from each input image using a pre-trained vision foundation model (e.g., DINOv3~\cite{dinov3}). These features encode view-invariant correspondences and structural regularities across images. We integrate the semantic features into the reconstruction pipeline to guide Gaussian prediction, encouraging globally consistent structure and coherent appearance across views. This design improves robustness to challenging scenarios such as large viewpoint changes, leading to better generalization on unseen scenes.

\noindent\textbf{Texture Encoder.}
Complementary to the semantic and geometry encoders, we introduce a texture encoder to capture high-frequency appearance details directly from input images. Unlike semantic features that encode high-level abstractions, appearance is inherently pixel-aligned and requires preserving local spatial fidelity. 
Specifically, the texture encoder employs lightweight convolutional layers to extract high-resolution features from each source view. 
These features are maintained in the image space and retain fine-grained color and texture information. We then inject the resulting features into the 3D feature stream through a pixel-aligned feature injection mechanism, establishing a direct correspondence between 2D observations and Gaussian attributes. 
By providing explicit appearance cues to the depth and Gaussian heads, this design enables accurate modeling of fine textures and consistent colors.

\begin{figure*}[t]
  \centering
  \includegraphics[width=\linewidth]{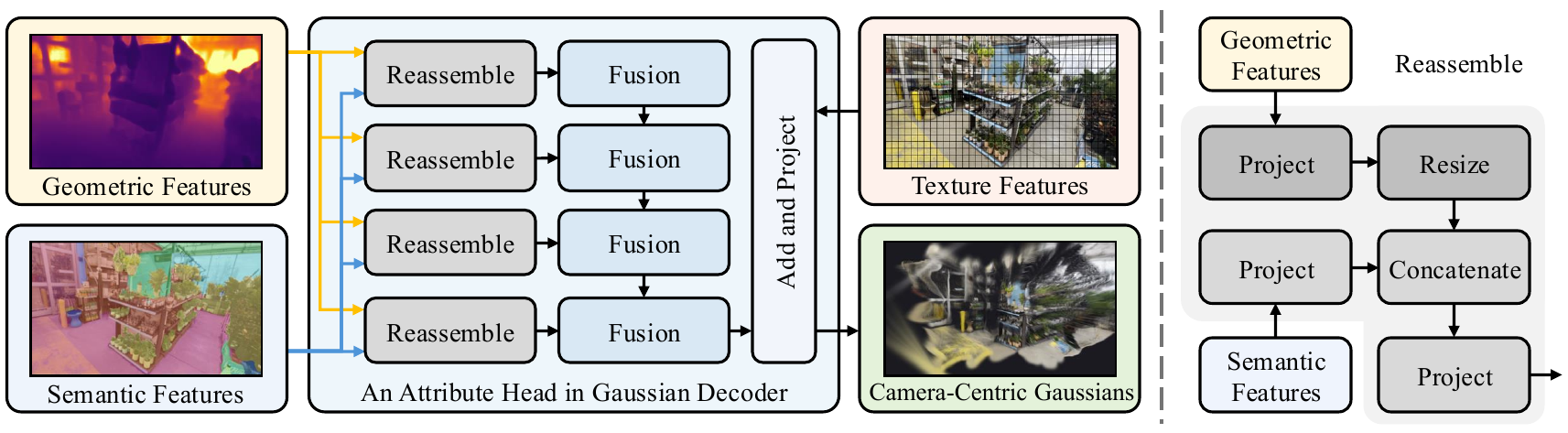}
  \caption{\textbf{Detailed architecture of the proposed Gaussian decoder head.} By modifying the standard DPT~\cite{dpt} heads, our decoder effectively integrates a structured representation to predict camera-centric Gaussian attributes. As detailed on the right, the redesigned \textit{Reassembling} block spatially aligns and concatenates geometric and semantic features, using projection layers to maintain channel dimensions. After processing through cascaded \textit{Reassembling} and \textit{Fusion} blocks, high-resolution texture features are injected at a late stage to explicitly preserve fine visual details before yielding the final output.}
  \label{fig:decoder}
\end{figure*}

\noindent\textbf{Gaussian Prediction.} The Gaussian prediction stage consists of a Gaussian decoder and a subsequent coordinate transformation. Initially, the Gaussian decoder predicts the attributes of Gaussian primitives in the local camera coordinate system of each source view, leveraging the structured representations. Subsequently, these camera-centric Gaussians are transformed into a unified world coordinate system based on the estimated camera parameters.

As illustrated in Fig.~\ref{fig:decoder}, we modify the standard DPT~\cite{dpt} heads to effectively integrate multi-scale semantic features and high-resolution texture cues. Specifically, the \textit{Reassembling} block is redesigned to align and concatenate geometric and semantic features, which are then projected to preserve the original DPT channel dimensions. The detailed structure of the \textit{Reassembling} block is shown on the right side of Fig.~\ref{fig:decoder}. After passing through cascaded \textit{Reassembling} and \textit{Fusion} blocks, high-resolution texture features are injected just before the final prediction layer to produce the Gaussian attributes. This late-stage integration explicitly preserves fine visual details and reduces texture degradation.

Finally, to construct a coherent 3D scene, the camera-centric predictions must be globally aligned. The decoder outputs the spatial position (parameterized by depth along the camera's z-axis) and the rotation quaternion defined in the local camera coordinate system for each Gaussian primitive. Consequently, a rigorous coordinate transformation step is applied to map these camera-space attributes into a unified world coordinate framework. This transformation utilizes the intrinsic and extrinsic  parameters derived from the camera decoder, ensuring strict spatial consistency across all source views.

\subsection{Camera Alignment for Leakage-Free Training}
\label{sec:cam_ali}

\begin{figure*}[t]
  \centering
  \includegraphics[width=\linewidth]{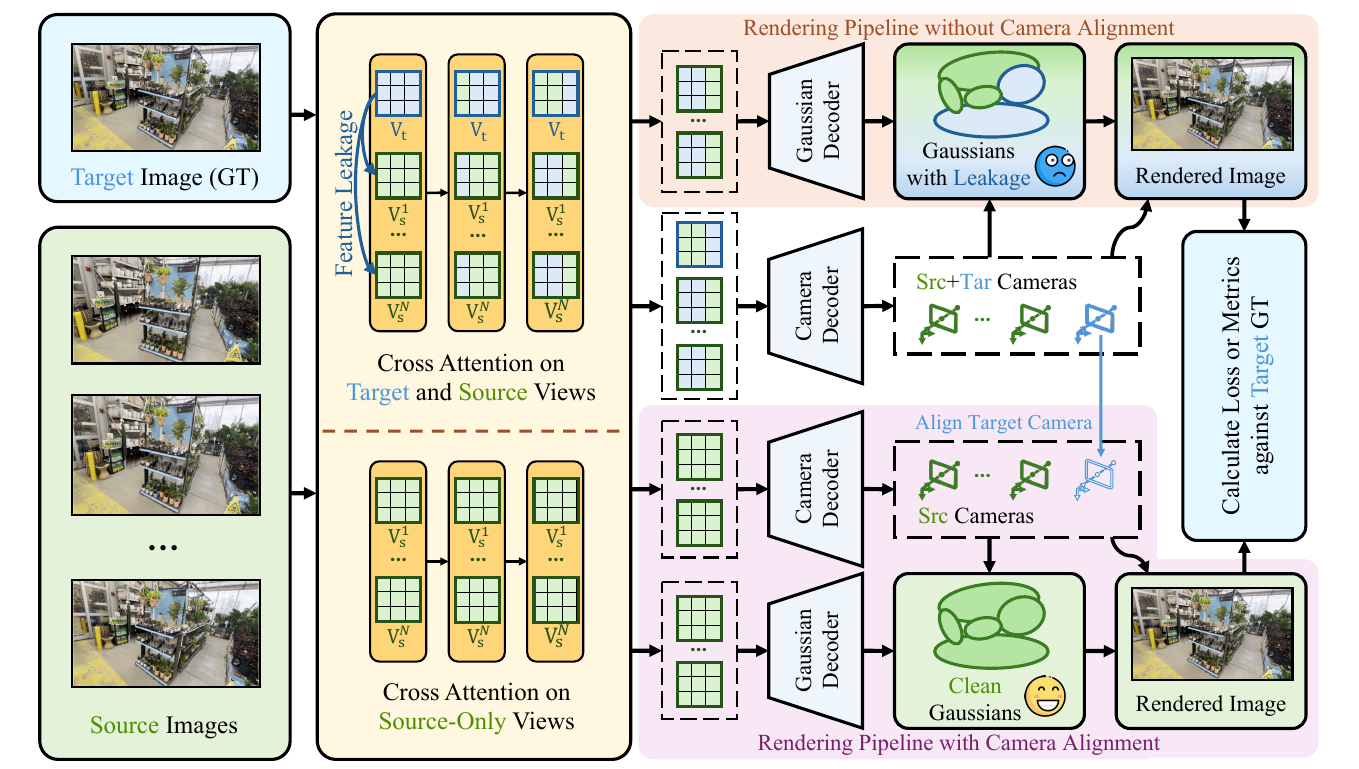}
  \caption{\textbf{Rendering pipeline with or without camera alignment.} Due to the cross-attention mechanism in the geometry extractor, target image information inevitably leaks into the source features. In contrast, our camera alignment strategy allows the Gaussian decoder to utilize pure features extracted from source-only views, thereby preventing feature leakage.}
  \label{fig:pipeline_wo_ca}
\end{figure*}

A key challenge arises during training: the target and source camera extrinsics must be established in a unified coordinate system. Simply incorporating target images into the source set to unify camera parameters and then separating their features for Gaussian prediction may introduce information leakage, which compromises generalization. As shown in Fig. \ref{fig:pipeline_wo_ca}, due to the cross-attention mechanism in the geometry extractor, target image information inevitably leaks into the source features. This leakage grants the Gaussian decoder implicit access to the target view's ground truth, compromising training supervision and evaluation fairness. To address this problem, we propose a parallel-stream camera pose alignment strategy that effectively isolates source and target features, enabling robust training and a fair evaluation within the existing framework. 

Our approach processes two image sets in parallel through the geometry feature extractor: a mixed set containing both source and target views, and a source-only set. The camera head then decodes the camera parameters for each set. Initially, these two sets of parameters reside in disparate world coordinate systems. We therefore align the target cameras from the mixed set to the coordinate system of the source-only set, using the source cameras as anchors. This process involves simultaneous optimization of rotation and translation.

\noindent\textbf{Rotation Alignment.}
To minimize computational overhead while preserving precision, we directly align the rotation quaternions predicted by the camera head. 
Let $q^{A}_{\mathrm{s},i}$ and $q^{B}_{\mathrm{s},i}$ denote the rotation quaternions of the $i$-th source view predicted from the mixed set $A$ and the source-only set $B$, respectively. 
We seek a unit quaternion $\Delta q$ that aligns $q^{A}_{\mathrm{s},i}$ with $q^{B}_{\mathrm{s},i}$ for all source views $i$.
This leads to the following constrained optimization problem:
\begin{equation}
\Delta q^{*} =
\arg\max_{\|\Delta q\|=1}
\sum_{i=1}^{n}
\mathrm{Re}\!\left(\Delta q\, q^{A}_{\mathrm{s},i}\, (q^{B}_{\mathrm{s},i})^{-1}\right).
\end{equation}
where $\mathrm{Re}(\cdot)$ extracts the real part of a quaternion and $n$ is the number of source views. 
We solve this problem using a Lagrange multiplier formulation.

\noindent\textbf{Translation Alignment.}
For the translation vectors $\tau^{A}_{\mathrm{s},i}$ and $\tau^{B}_{\mathrm{s},i}$, 
we seek a scale $\lambda \in \mathbb{R}$ and an offset $\Delta \tau \in \mathbb{R}^{3}$ 
that align $\lambda\, \tau^{A}_{\mathrm{s},i} + \Delta \tau$ with $\tau^{B}_{\mathrm{s},i}$ for all source views $i$. 
This can be formulated as a least-squares problem:
\begin{equation}
(\lambda^{*}, \Delta \tau^{*}) 
=
\arg\min_{\lambda,\Delta \tau}
\sum_{i=1}^{n}
\left\|
\lambda\, \tau^{A}_{\mathrm{s},i} + \Delta \tau - \tau^{B}_{\mathrm{s},i}
\right\|_{2}^{2}.
\end{equation}

Finally, we transform the target-view poses from the mixed set into the Gaussian representation coordinate system (i.e., the source-only set) as 
$q^{B}_{\mathrm{t},j} = \Delta q^{*} q^{A}_{\mathrm{t},j}$ and 
$\tau^{B}_{\mathrm{t},j} = \lambda^{*} \tau^{A}_{\mathrm{t},j} + \Delta \tau^{*}$,
where $(q^{A}_{\mathrm{t},j}, \tau^{A}_{\mathrm{t},j})$ and $(q^{B}_{\mathrm{t},j}, \tau^{B}_{\mathrm{t},j})$ denote the camera poses of the $j$-th target view before and after alignment, respectively. 
As illustrated in Fig.~\ref{fig:overivew}, this strategy ensures that the modules responsible for predicting Gaussian attributes never access features from the target views. Meanwhile, it aligns the target views into a consistent world coordinate system, which is crucial for model generalization and fair evaluation.

%% file: sections/04_experiment.tex

\begin{table}[t]
\centering
\caption{\textbf{Comparison with representational reconstruction methods.} The best and second-best results of methods without any camera parameters are highlighted in \textbf{bold} and \underline{underlined}, respectively.  Note that we evaluate all metrics without any post optimization in this table. (``Weak'' denotes only relying on intrinsics and ``No'' denotes methods without any camera parameters.)}
\label{tab:maincomparison}
\resizebox{\linewidth}{!}{
\begin{threeparttable}
\begin{tabular}{@{}lcccccccccc@{}}
\toprule
\multirow{2}{*}{Method} & \multirow{2}{*}{\makecell{Camera \\ Condition}} & \multicolumn{3}{c}{DL3DV~\cite{dl3dv}} & \multicolumn{3}{c}{ACID~\cite{acid}} & \multicolumn{3}{c}{RealEstate10K~\cite{re10k}} \\ \cline{3-11} 
&    & PSNR$\uparrow$   & SSIM$\uparrow$   & LPIPS$\downarrow$  & PSNR$\uparrow$    & SSIM$\uparrow$   & LPIPS$\downarrow$ & PSNR$\uparrow$       & SSIM$\uparrow$      & LPIPS$\downarrow$    \\ \midrule
PF3plat~\cite{pf3plat}      & Weak & 22.579 & 0.731 & 0.186 & 23.401 & 0.729 & 0.279 & 22.766 & 0.784 & 0.190 \\
SelfSplat~\cite{selfsplat}  & Weak & 18.017 & 0.511 & 0.366 & 23.528 & 0.729 & 0.279 & 22.542 & 0.781 & 0.229 \\
NoPoSplat~\cite{noposplat}  & Weak & 25.592 & 0.838 & 0.104 & 25.521 & 0.750 & 0.185 & 24.724 & 0.818 & 0.145 \\
FLARE$^{\ast}$~\cite{flare}          & Weak & 23.441 & 0.757 & 0.126 & 24.365 & 0.702 & 0.200 & 23.611 & 0.777 & 0.155 \\
\midrule
Splatt3R~\cite{splatt3r}    & No  & 15.936 & 0.391 & 0.408 & 11.000 & 0.307 & 0.582 & 12.092 & 0.364 & 0.540  \\
RayZer~\cite{rayzer}        & No  & 19.802 & 0.527 & 0.286 & -      & -     & -     & -      & -     & -     \\
AnySplat~\cite{anysplat}    & No  & \underline{22.377} & \underline{0.716} & \underline{0.150}  & \underline{22.433} & \underline{0.651} & \underline{0.237} & \underline{20.521} & \underline{0.686} & \underline{0.212} \\
Depth Anything 3~\cite{da3} & No  & 20.715 & 0.615 & 0.226 & 20.482 & 0.588 & 0.346 & 18.769 & 0.613 & 0.312 \\
StructSplat (Ours)       & No  &\textbf{28.045} & \textbf{0.888} & \textbf{0.091} & \textbf{24.372} & \textbf{0.712} & \textbf{0.219} & \textbf{22.240} & \textbf{0.729} & \textbf{0.201} \\
\bottomrule
\end{tabular}
\begin{tablenotes}
\footnotesize
\item[$\ast$] FLARE~\cite{flare} natively utilizes camera intrinsics as conditional input for \textit{novel-view-synthesis} subtasks to align with the experimental settings of NoPoSplat~\cite{noposplat}.
\end{tablenotes}
\end{threeparttable}
}
\end{table}
\section{Experiments}
In this section, we first describe the experimental setup, including datasets and baseline methods. We then present the main comparisons with state-of-the-arts. 
Finally, we conduct ablation studies and  further analyses to assess the contribution of each component in our framework.
\subsection{Experiment Setting}
\label{sec:experiment_setting}

\noindent\textbf{Datasets and Evaluation.}
We evaluate our method on three large-scale datasets: DL3DV~\cite{dl3dv}, RealEstate10K~\cite{re10k}, and ACID~\cite{acid}. 
For DL3DV, we follow prior works~\cite{depthsplat, pf3plat} and use the official benchmark split as the test set. 
For RealEstate10K and ACID, we use the model trained on DL3DV for cross-dataset evaluation.
To ensure a fair evaluation, we carefully remove overlapping and anomalous samples from the training set. Since baseline methods render images at different resolutions, we uniformly resize and center-crop all outputs and ground-truth images to $256 \times 256$. 
We report PSNR, SSIM~\cite{ssim}, and LPIPS~\cite{lpips} as evaluation metrics. 

\noindent\textbf{Baseline Methods.}
We comprehensively benchmark our approach against a variety of state-of-the-art baselines. Based on their input requirements, these baselines can be categorized into two main groups: 
1) pose-free but intrinsics-dependent methods, which operate without camera poses but still rely on known camera intrinsics, including PF3plat \cite{pf3plat}, SelfSplat \cite{selfsplat}, NoPoSplat \cite{noposplat}, and FLARE \cite{flare}; 
2) entirely camera-parameter-free methods, comprising Splatt3R \cite{splatt3r}, RayZer \cite{rayzer}, AnySplat \cite{anysplat}, and Depth Anything 3 \cite{da3}. 
Notably, to align with the experimental settings of NoPoSplat~\cite{noposplat}, FLARE~\cite{flare} natively utilizes camera intrinsics as conditional input for novel-view-synthesis subtasks.

\noindent\textbf{Implementation.}
To optimize training efficiency and alleviate GPU memory overhead, we incorporate bfloat16 (BF16) mixed-precision training \cite{bf16} alongside the DeepSpeed optimization library \cite{zero1, zero2, zero3}. 
During the training phase, we employ the Warmup-Stable-Decay (WSD) learning rate scheduler \cite{wsd}, which ensures stable convergence and provides flexibility for continuous model retraining. 
To fully leverage the available data and enforce multi-view consistency, we jointly render both the source and target views to compute the training loss.
For final evaluation, all quantitative metrics are computed exclusively on the target views.

\subsection{Comparisons with State-of-the-Art Methods}

\begin{figure}[t]
  \centering
  \begin{overpic}[width=\linewidth, trim = 0 0 0 10, clip]{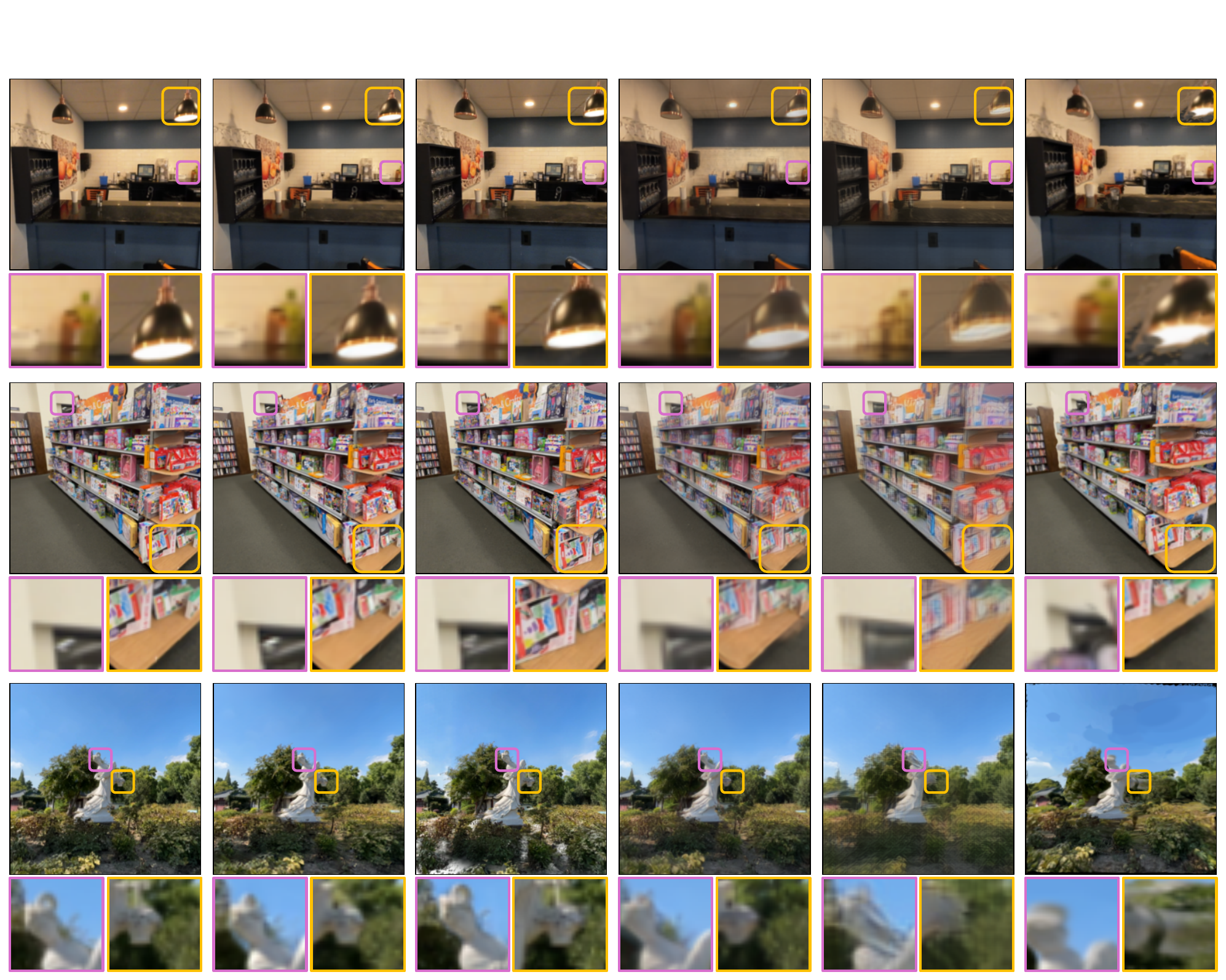}
    \put(7, 74.5){\textcolor{black}{GT}}
    \put(22.5 , 74.5){\textcolor{black}{Ours}}
    \put(34, 74.5){\textcolor{black}{AnySplat~\cite{anysplat}}}
    \put(52.5, 74.5){\textcolor{black}{DA 3~\cite{da3}}}
    \put(68, 74.5){\textcolor{black}{RayZer~\cite{rayzer}}}
    \put(84, 74.5){\textcolor{black}{Splatt3R~\cite{splatt3r}}}
  \end{overpic}
  \caption{\textbf{Visual comparison of novel view synthesis on DL3DV}~\cite{dl3dv} \textbf{dataset.}}
  \label{fig:vis_cmp}
\end{figure}
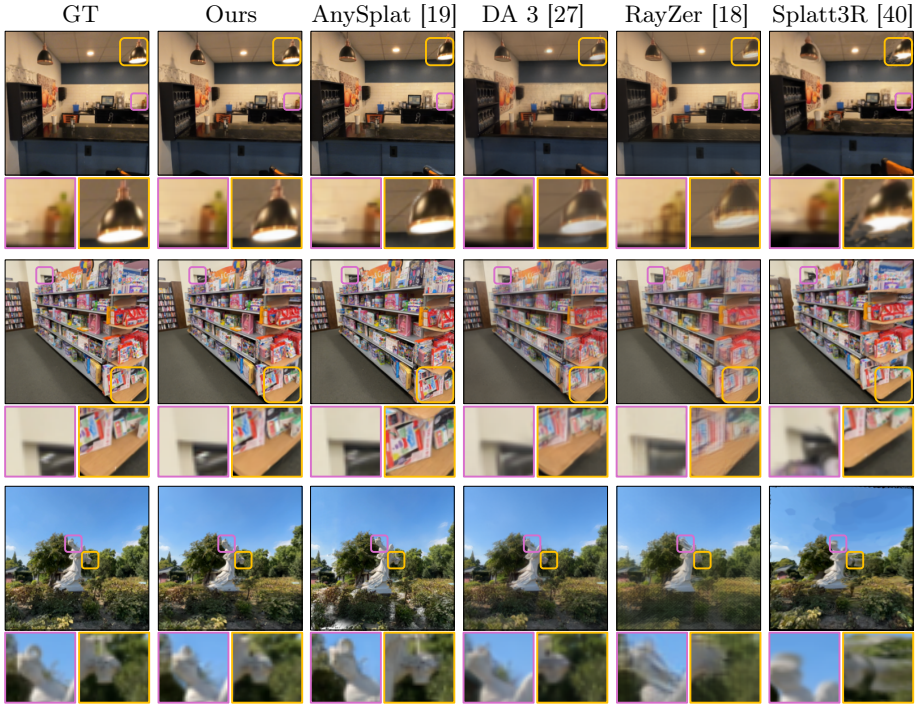

\noindent\textbf{Quantitative Comparison.}
The quantitative comparison with state-of-the-art methods is reported in Table~\ref{tab:maincomparison}. 
Overall, our method consistently achieves the best performance across all datasets and metrics among approaches that do not rely on camera parameters. On DL3DV, our method significantly outperforms prior works, achieving a PSNR of 28.045, which surpasses the best competing method AnySplat (22.377) by a large margin of +5.67 dB. Similar gains are observed in SSIM (0.888 vs. 0.716) and LPIPS (0.091 vs. 0.150), demonstrating substantial improvements in both structural fidelity and perceptual quality.

On ACID, we use the models trained on DL3DV as a cross-dataset evaluation. Our method achieves the best performance among camera-free approaches, with a PSNR of 24.372, outperforming AnySplat (22.433) by +1.939 dB. 
Consistent improvements are also observed in SSIM and LPIPS, indicating better structural fidelity and perceptual quality. 
Given the challenging nature of ACID, which involves complex camera motions and diverse scene layouts, these results demonstrate the robustness of our method in handling uncalibrated and geometrically challenging scenarios.

On RealEstate10K, which contains more complex real-world scenes with diverse layouts and watermark artifacts, we use the models trained on DL3DV as a cross-dataset evaluation. Our method continues to achieve the best performance among methods without camera parameters. 
Specifically, it achieves a PSNR of 22.240, outperforming AnySplat (20.521) by +1.719 dB, and also obtains higher SSIM (0.729 vs. 0.686) and lower LPIPS (0.201 vs. 0.212). 
Notably, despite the increased difficulty of this dataset, our method remains competitive with approaches that rely on camera intrinsics, such as PF3plat (22.766 PSNR), highlighting the robustness of our camera-free design.

Importantly, even compared with methods that rely on camera intrinsics (``Weak'' setting), our approach achieves competitive or superior results. 
For instance, on DL3DV, our method (28.045 PSNR) clearly surpasses NoPoSplat (25.592), highlighting the effectiveness of our fully camera-free design.

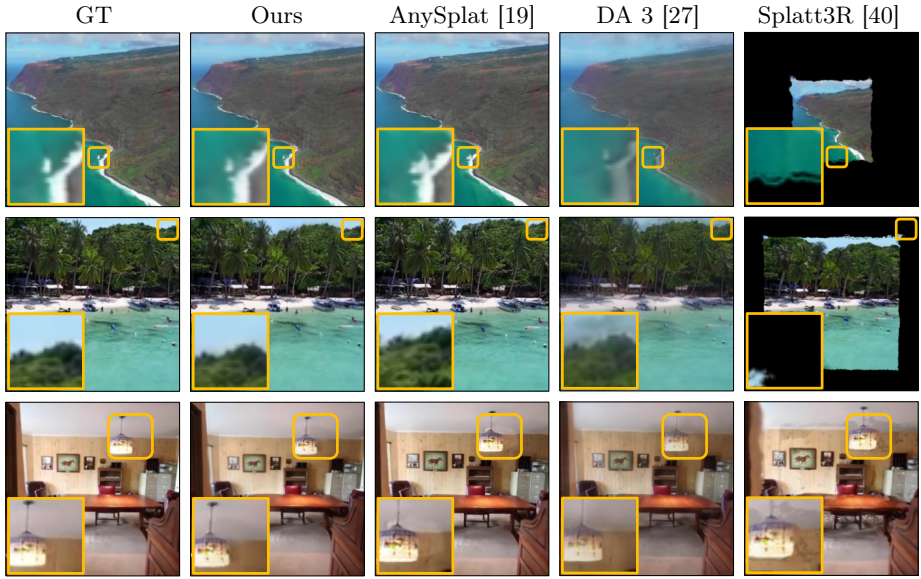
\begin{figure}[t]
  \centering
  \begin{overpic}[width=\linewidth, trim = 0 0 0 10, clip]{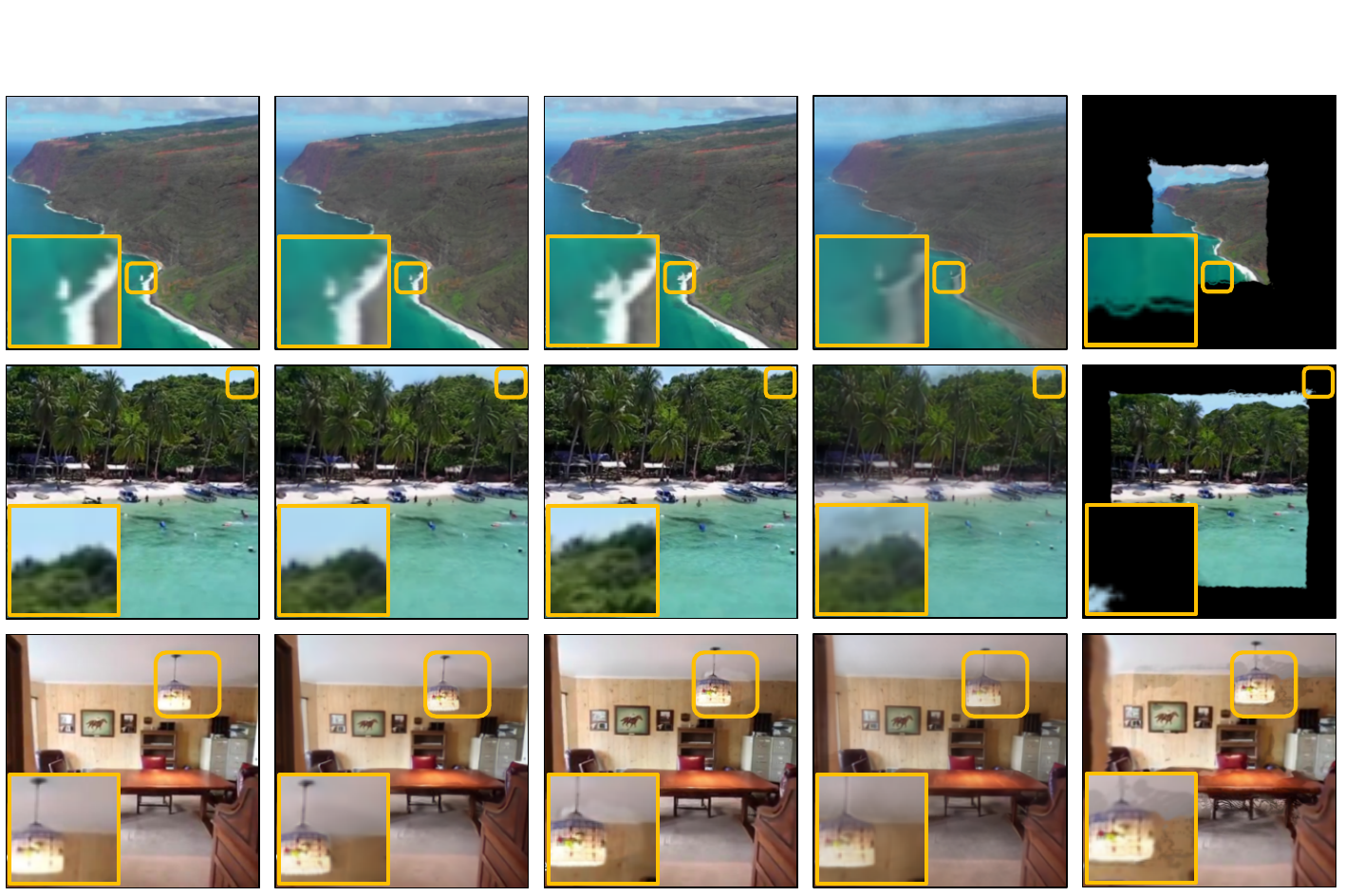}
    \put(8, 60.5){\textcolor{black}{GT}}
    \put(27 , 60.5){\textcolor{black}{Ours}}
    \put(42, 60.5){\textcolor{black}{AnySplat~\cite{anysplat}}}
    \put(64.5, 60.5){\textcolor{black}{DA 3\cite{da3}}}
    \put(82, 60.5){\textcolor{black}{Splatt3R~\cite{splatt3r}}}
  \end{overpic}
  \caption{\textbf{Visual comparison of novel view synthesis on RealEstate10K}~\cite{re10k} \textbf{and ACID}~\cite{acid} \textbf{datasets.}}
  \label{fig:vis_acid}
\end{figure}

\noindent\textbf{Qualitative Comparison.} 
Fig.~\ref{fig:vis_cmp} presents qualitative results on the DL3DV dataset~\cite{dl3dv}. 
As shown in the zoomed-in regions, our approach recovers fine-grained details and produces sharper edges, whereas baseline methods tend to generate blurry or distorted results. 
Our method also preserves more consistent textures and structural boundaries across views, leading to more faithful scene reconstruction. Additionally, visual comparisons on the RealEstate10K~\cite{re10k} and ACID~\cite{acid} datasets are provided in Fig.~\ref{fig:vis_acid}. 
Despite the increased scene complexity and viewpoint variations in these datasets, our method reconstructs coherent 3D structures and produces photorealistic views that align closely with the Ground Truth.

\begin{table}[t]
\caption{\textbf{Comparison for multi-view reconstruction on DL3DV}~\cite{dl3dv}\textbf{.} The best and second-best results of methods without any camera parameters are highlighted in \textbf{bold} and \underline{underlined}, respectively.}
\label{tab:multi_views}
\setlength{\tabcolsep}{1.5pt}
\resizebox{\linewidth}{!}{
\begin{tabular}{@{}lccccccccc@{}}
\toprule
\multirow{2}{*}{Method} &\multicolumn{3}{c}{2 Source Views} & \multicolumn{3}{c}{4 Source Views} & \multicolumn{3}{c}{6 Source Views} \\ \cmidrule(l){2-10} 
                        & PSNR$\uparrow$      & SSIM$\uparrow$     & LPIPS$\downarrow$ & PSNR$\uparrow$       & SSIM$\uparrow$      & LPIPS$\downarrow$     & PSNR$\uparrow$    &SSIM$\uparrow$      & LPIPS$\downarrow$    \\ \midrule
RayZer~\cite{rayzer}     & 19.802 & 0.527 & 0.286 & \underline{27.888}  & \underline{0.868} & 0.140 & \underline{28.851}  & \underline{0.889} & 0.128 \\
AnySplat~\cite{anysplat} & \underline{22.377} & \underline{0.716} & \underline{0.150}  & 24.499 & 0.769 & \underline{0.112} & 26.529 & 0.841 & \underline{0.085}  \\
Depth Anything 3~\cite{da3}  & 20.715 & 0.615 & 0.226 & 20.486 & 0.601 & 0.232 & 20.566 & 0.602 & 0.232    \\
StructSplat (Ours)         & \textbf{28.045}   & \textbf{0.888}  & \textbf{0.091} & \textbf{30.415}  & \textbf{0.931} & \textbf{0.061} & \textbf{31.137} & \textbf{0.942} & \textbf{0.057}     \\ \bottomrule
\end{tabular}}
\end{table}

Furthermore, we investigate the multi-view reconstruction capability of our method on DL3DV~\cite{dl3dv} using 2, 4 and 6 source views. As shown in Table \ref{tab:multi_views}, our method consistently outperforms other methods across different numbers of input views, demonstrating its capability for high-quality 3D reconstruction even with multiple source views.

\subsection{Ablation Studies and Further Analyses}

\begin{table}[t]
\RawFloats
\centering

\begin{minipage}{0.52\linewidth}
\centering
\captionof{table}{\textbf{Comparison for ablation of Semantic Features and Texture Features on DL3DV}~\cite{dl3dv}\textbf{.} The best results are highlighted in \textbf{bold}.}
\label{tab:ab_of_tf_sf}
\begin{tabular}{@{}lccc@{}}
\toprule
Experiment Setting   & PSNR$\uparrow$  & SSIM$\uparrow$ & LPIPS$\downarrow$ \\
\midrule
Geometric Features    & 20.610  & 0.622  & 0.343 \\
+ Semantic Features   & 26.236  & 0.848  & 0.151 \\
+ Texture Features    & \textbf{28.045} & \textbf{0.888} & \textbf{0.091} \\
\bottomrule
\end{tabular}
\end{minipage}
\hfill
\begin{minipage}{0.46\linewidth}
\centering
\captionof{table}{\textbf{Impact of pose post-optimization on DL3DV.} The best and second-best results are highlighted in \textbf{bold} and \underline{underlined}, respectively.}
\label{tab:pose-post-opt}
\begin{tabular}{@{}lccc@{}}
\toprule
Method   & PSNR$\uparrow$ & SSIM$\uparrow$ & LPIPS$\downarrow$ \\
\midrule
NoPoSplat~\cite{noposplat} & \underline{27.243} & \underline{0.877} & \underline{0.093} \\
FLARE~\cite{flare} & 26.916 & 0.865 & 0.099 \\
Ours & \textbf{29.287} & \textbf{0.911} & \textbf{0.085} \\
\bottomrule
\end{tabular}
\end{minipage}
\vspace{-0.3cm} 
\end{table}

\noindent\textbf{Impact of Semantic Encoder.}
To evaluate the role of semantic priors, we first augment a baseline model that uses only \textit{Geometric Features} with \textit{Semantic Features}. 
As shown in Table~\ref{tab:ab_of_tf_sf}, introducing semantic features leads to a substantial performance gain, improving PSNR from 20.610 to 26.236. This significant improvement indicates that high-level semantic context plays a critical role in resolving structural ambiguities and enhancing global consistency in complex scenes.

\noindent\textbf{Impact of Texture Encoder.}
We further analyze the contribution of pixel-aligned appearance modeling by adding \textit{Texture Features} on top of geometric and semantic features. This results in additional performance improvements, achieving the best results across all metrics (PSNR 28.045). These gains demonstrate that texture features are essential for recovering fine-grained details and accurate colors, complementing the structural guidance provided by geometry and semantics.

\noindent\textbf{Impact of Pose Post-optimization.} We further evaluate the effect of camera pose refinement~\cite{noposplat} by optimizing the target extrinsics for a few iterations.
As shown in Table~\ref{tab:pose-post-opt}, our method continues to outperform NoPoSplat~\cite{noposplat} and FLARE~\cite{flare}, indicating that the advantage of our structured representation persists even when additional pose refinement is applied.

\noindent\textbf{Impact of Camera Alignment.}
To ensure robust feed-forward reconstruction, we incorporate a camera alignment strategy to enforce a consistent coordinate system across views. AnySplat~\cite{anysplat} also employs a simpler strategy that aligns the translation scale during evaluation. We compare the Absolute Trajectory Error (ATE)~\cite{zhang2018tutorial} of target camera poses obtained using these two alignment strategies in Table~\ref{tab:cp_of_ca}. Our StructSplat achieves lower errors in both rotation and translation.
As shown in Table~\ref{tab:ab_of_ca}, removing this component leads to noticeable performance degradation, demonstrating its critical role in maintaining reconstruction fidelity. 
The alignment mechanism effectively prevents information leakage between source and target views, which is crucial for learning a more generalizable representation. 
As a result, the model achieves stronger generalization ability to unseen scenes, particularly in cases with complex geometry.

\begin{table}[t]
\RawFloats
\centering
\begin{minipage}{0.46\linewidth}
\centering
\caption{\textbf{Comparison for different Camera Alignment strategies on DL3DV}~\cite{dl3dv}\textbf{.} The best results are highlighted in \textbf{bold}.}
\label{tab:cp_of_ca}
\begin{tabular}{@{}lcc@{}}
\toprule
Method        & ATE$_\text{rot} \downarrow$  & ATE$_\text{pos} \downarrow$    \\ \midrule
AnySplat~\cite{anysplat} & 0.0866  & 0.1003                    \\
StructSplat (Ours) & \textbf{0.0016}  & \textbf{0.0093}        \\ \bottomrule
\end{tabular}

\end{minipage}
\hfill
\begin{minipage}{0.51\linewidth}
\centering
\caption{\textbf{Rendering comparison for ablation of our Camera Alignment strategy on DL3DV}~\cite{dl3dv}\textbf{.} The best results are highlighted in \textbf{bold}.} 
\label{tab:ab_of_ca}
\begin{tabular}{@{}lccc@{}}
\toprule
Train Setting        & PSNR$\uparrow$  & SSIM$\uparrow$ & LPIPS$\downarrow$                    \\ \midrule
w/o CA & 27.338  & 0.879  & 0.097                     \\
StructSplat (Ours) & \textbf{28.045}  & \textbf{0.888}  & \textbf{0.091}                     \\ \bottomrule
\end{tabular}
\end{minipage}
\vspace{-0.3cm} 
\end{table}

\noindent\textbf{Real-world Applications.}
We further evaluate our method on real-world, in-the-wild image collections without camera calibration, including Internet videos and handheld videos captured in casual settings. 
Despite the absence of camera parameters, our method reconstructs coherent 3D structures and produces consistent novel views under challenging conditions such as complex scene layouts.

%% file: sections/05_conclusion.tex
\section{Conclusion}

In this paper, we presented StructSplat, a feed-forward and generalizable 3D Gaussian reconstruction framework that operates on uncalibrated images without requiring camera parameters. We addressed key limitations of existing methods by introducing a structured representation that models geometry, appearance, and semantics in their native spaces, together with a pixel-aligned feature injection mechanism and a camera alignment strategy to prevent information leakage. Extensive experiments demonstrate that our method achieves superior performance over prior approaches. Despite these improvements, challenges remain under extremely sparse views or severe occlusions, and modeling complex view-dependent effects and lighting remains difficult. Future work will focus on more robust geometry priors and improved appearance modeling, as well as extending the framework to dynamic scenes and large-scale environments.

%% file: sections/06_acknowledgements.tex
\section*{Acknowledgements}

This work was supported by the National Natural Science Foundation of China (Grant No. 62506063 and No. 62306085), the Guangdong Basic and Applied Basic Research Foundation (Grant No. 2024A1515110178), the Start-up Funding (Grant No. YJKY230070), the Shenzhen College Stability Support Plan (Grant No. GXWD20231130151329002) and the Shenzhen Science and Technology Program (Grant No. KQTD20240729102207002). Computational resources were provided by the Songshan Lake High Performance Computing Center (SSL-HPC) at Great Bay University. This work was also supported by the Guangdong Research Team for Communication and Sensing Integrated with Intelligent Computing (Project No. 2024KCXTD047).

%% file: sections/supplementary.tex
\newpage
\appendix
\section{Implementation and Architecture Details}

\noindent\textbf{Gaussian Activations.} As summarized in Table~\ref{tab:activation}, the output activations of different attribute heads in the Gaussian decoder are selected according to the statistical distributions of their targets. For the color branch, which is supervised by ground-truth values in $[0,1]$, we adopt a linear activation to prevent gradient saturation. Consequently, no explicit sigmoid is required, as the regression loss naturally enforces valid-range predictions.

\begin{table}[htbp]
\centering
\caption{\textbf{Output activations and channels of different attribute heads in Gaussian decoder.}} 
\label{tab:activation}
\begin{tabular*}{0.6\textwidth}{@{\extracolsep{\fill}}llc}
\toprule
Attribute    & Activation                  & Channels \\ \midrule
Depth        & $y=2\mathrm{Softplus}(x;\beta=2\ln2)$ & 1        \\
Opacity      & $y=\mathrm{Sigmoid}(x-2)$            & 1        \\
Color        & $y=0.25x$                   & 3        \\
Scale        & $y=\mathrm{Softplus}(x;\beta=\ln2)$   & 3        \\
Rotation     & $y=\frac{x}{\|x\|_2}$        & 4        \\ \bottomrule
\end{tabular*}
\end{table}

\noindent\textbf{Training Loss.}
The joint training of our Gaussian representation is guided by a multi-term loss function, where each component emphasizes a distinct aspect of rendering quality. The overall objective is formulated as:
\begin{align}
\nonumber \mathcal{L}_\mathrm{Gaussian}(\mathcal{I},\widehat{\mathcal{I}})
&=\sum_{i=1}^{N_\mathrm{src}+N_\mathrm{tar}}
\Big[
w_1\,\mathrm{MSE}(I_i,\widehat{I}_i) \\
\nonumber
&\quad + w_2\left(1-\mathrm{SSIM}(I_i,\widehat{I}_i)\right) \\
&\quad + w_3\,\mathrm{LPIPS}(I_i,\widehat{I}_i)
\Big].
\end{align}

This objective combines the Mean Squared Error (MSE) for pixel-level fidelity, the Structural Similarity Index Measure (SSIM)~\cite{ssim} to preserve structural consistency, and the Learned Perceptual Image Patch Similarity (LPIPS)~\cite{lpips} to improve perceptual realism.

During training, the ground-truth set $\mathcal{I}=\{I_i\}_{i=1}^{N_\mathrm{src}+N_\mathrm{tar}}$ and the rendered set $\widehat{\mathcal{I}}=\{\widehat{I}_i\}_{i=1}^{N_\mathrm{src}+N_\mathrm{tar}}$ include both $N_\mathrm{src}$ source views and $N_\mathrm{tar}$ target views. For evaluation, all quantitative metrics are computed exclusively on the target views.

\noindent\textbf{Solution for Camera Alignment.}
We formulate the \textbf{Camera Alignment} as two optimization problems:
\begin{align}
& \Delta q^{*} =
\arg\max_{\|\Delta q\|=1}
\sum_{i=1}^{n}
\mathrm{Re}\!\left(\Delta q\, q^{A}_{\mathrm{s},i}\,(q^{B}_{\mathrm{s},i})^{-1}\right),  \\
& (\lambda^{*}, \Delta \tau^{*}) 
=
\arg\min_{\lambda,\Delta \tau}
\sum_{i=1}^{n}
\left\|
\lambda\, \tau^{A}_{\mathrm{s},i} + \Delta \tau - \tau^{B}_{\mathrm{s},i}
\right\|_{2}^{2}.
\end{align}

For rotation, the solution derived via the Lagrange multiplier method is
\begin{equation}
\Delta q^{*}=
\frac{\sum_{i=1}^n q^{B}_{\mathrm{s},i}\,(q^{A}_{\mathrm{s},i})^{-1}}
{\left\|\sum_{i=1}^n q^{B}_{\mathrm{s},i}\,(q^{A}_{\mathrm{s},i})^{-1}\right\|}.
\end{equation}

For translation, setting the gradient to zero yields
\begin{align}
\lambda^{*} &=
\frac{
\sum_{i=1}^{n}\tau^{A}_{\mathrm{s},i}\cdot \tau^{B}_{\mathrm{s},i}
-
\frac{1}{n}
\sum_{i=1}^{n}\tau^{A}_{\mathrm{s},i}\cdot
\sum_{i=1}^{n}\tau^{B}_{\mathrm{s},i}
}{
\sum_{i=1}^{n}\left\|\tau^{A}_{\mathrm{s},i}\right\|_2^{2}
-
\frac{1}{n}
\left\|
\sum_{i=1}^{n}\tau^{A}_{\mathrm{s},i}
\right\|^{2}
}, \\
\Delta \tau^{*} &=
\frac{
\sum_{i=1}^{n}\tau^{B}_{\mathrm{s},i}
-
\lambda^{*}
\sum_{i=1}^{n}\tau^{A}_{\mathrm{s},i}
}{n}.
\end{align}

\noindent\textbf{Computation Details.}
Our StructSplat is trained using a single H100 GPU for two-views setting. The total training time is approximately 2.5 days, with a peak memory usage of 73 GB. In terms of model complexity and efficiency, our model contains 1.535 B parameters and requires 164.657 ms per inference pass.

\begin{table}[t]
\caption{\textbf{Comparison for larger-scale reconstruction on DL3DV}~\cite{dl3dv}\textbf{.} The best and second-best results of methods without any camera parameters are highlighted in \textbf{bold} and \underline{underlined}, respectively.}
\label{tab:larger_scene}
\setlength{\tabcolsep}{1.5pt}
\resizebox{\linewidth}{!}{
\begin{tabular}{@{}lcccccccccccc@{}}
\toprule
\multirow{2}{*}{Method} &\multicolumn{3}{c}{4 Source Views} & \multicolumn{3}{c}{6 Source Views} & \multicolumn{3}{c}{12 Source Views} & \multicolumn{3}{c}{18 Source Views} \\ \cmidrule(l){2-13} 
                        & PSNR$\uparrow$      & SSIM$\uparrow$     & LPIPS$\downarrow$ & PSNR$\uparrow$       & SSIM$\uparrow$      & LPIPS$\downarrow$     & PSNR$\uparrow$    &SSIM$\uparrow$      & LPIPS$\downarrow$  & PSNR$\uparrow$    &SSIM$\uparrow$      & LPIPS$\downarrow$  \\ \midrule
AnySplat~\cite{anysplat} & \underline{16.852} & \underline{0.446} & \underline{0.310}  & \underline{18.125} & \underline{0.496} & \underline{0.273} & \underline{17.887} & \underline{0.487} & \underline{0.300} & \underline{18.920} & \underline{0.539} & \underline{0.269} \\
Depth Anything 3~\cite{da3}  & 16.780 & 0.443 & 0.391  & 17.126 & 0.455 & 0.370 & 14.771 & 0.364 & 0.494 & 14.290 & 0.368 & 0.554 \\
StructSplat (Ours)  & \textbf{20.171}   & \textbf{0.692}  & \textbf{0.231} & \textbf{22.358}  & \textbf{0.749} & \textbf{0.187} & \textbf{22.173} & \textbf{0.731} & \textbf{0.205}  & \textbf{22.396} & \textbf{0.752} & \textbf{0.196}  \\ \bottomrule
\end{tabular}}
\end{table}

\begin{table}[h]
\caption{\textbf{Comparison for high resolutions on DL3DV}~\cite{dl3dv}\textbf{.} The best and second-best results of methods without any camera parameters are highlighted in \textbf{bold} and \underline{underlined}, respectively.}
\label{tab:high_res}
\setlength{\tabcolsep}{1.5pt}
\begin{tabular}{@{}lcccccc@{}}
\toprule
\multirow{2}{*}{Method} &\multicolumn{3}{c}{448$\times$448 Resolution} & \multicolumn{3}{c}{504$\times$504 Resolution}  \\ \cmidrule(l){2-7} 
                        & PSNR$\uparrow$      & SSIM$\uparrow$     & LPIPS$\downarrow$ & PSNR$\uparrow$       & SSIM$\uparrow$      & LPIPS$\downarrow$    \\ \midrule
AnySplat~\cite{anysplat} & \underline{20.870} & \underline{0.640} & \underline{0.218}  & - & - & - \\
Depth Anything 3~\cite{da3}  & 20.169 & 0.598 & 0.272  & \underline{19.745} & \underline{0.575} & \underline{0.295} \\
StructSplat (Ours)  & \textbf{27.073}   & \textbf{0.859}  & \textbf{0.126} & \textbf{26.886}  & \textbf{0.853} & \textbf{0.137} \\ \bottomrule
\end{tabular}
\end{table}

\begin{figure}[h]
  \centering
  \includegraphics[width=0.95\linewidth]{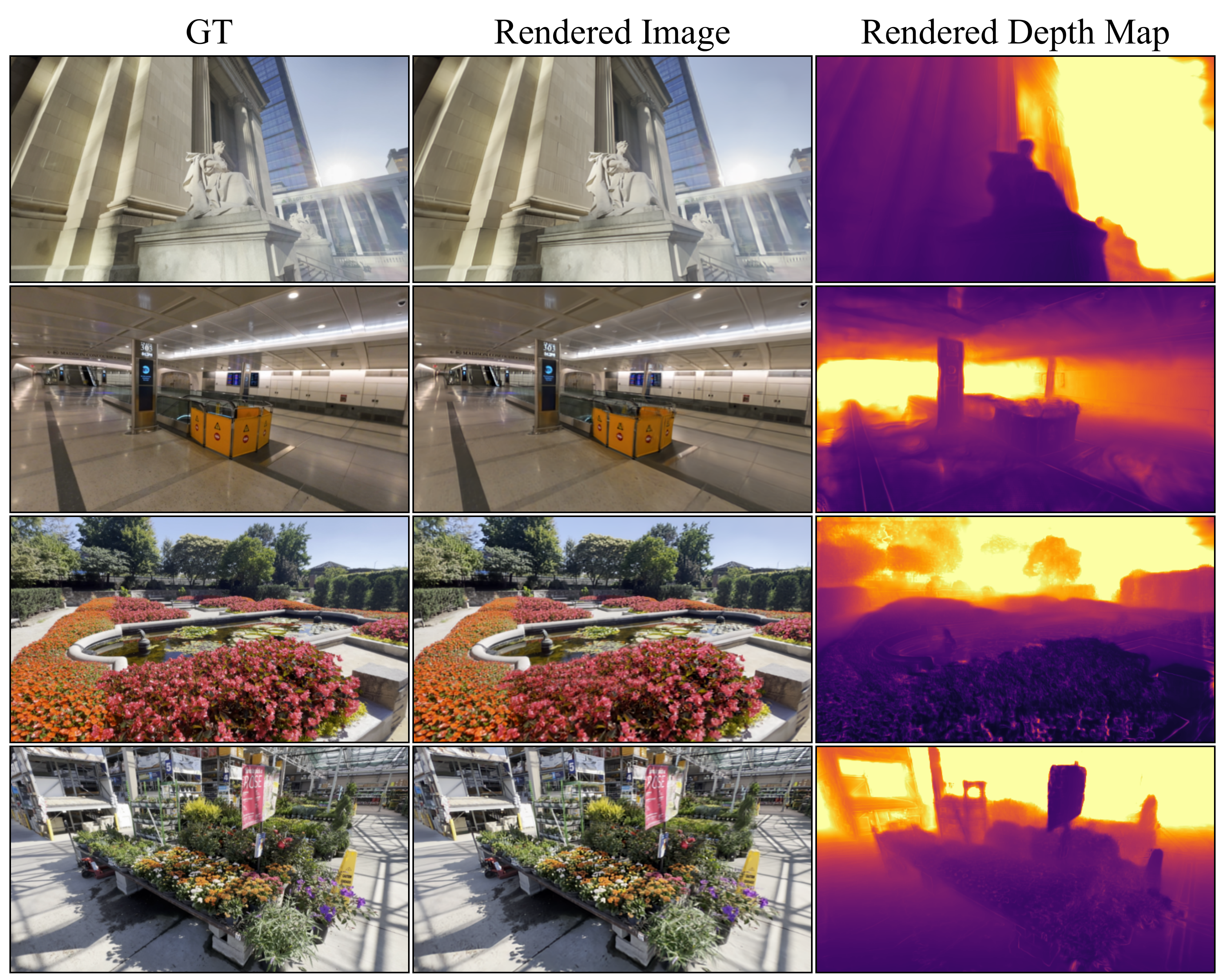}
  \caption{\textbf{Qualitative visualization of rendered novel views and depth maps.} From left to right: Ground-truth images, RGB images rendered by our method, and the corresponding rendered depth maps. Although our model is trained exclusively with 2D photometric and perceptual supervision from RGB images without any ground-truth depth data or explicit depth consistency losses, it successfully captures the underlying 3D geometry, yielding highly accurate and structurally consistent depth maps across diverse scenes.
  \vspace{-0.3cm}
  }
  \label{fig:depth}
\end{figure}

\begin{figure}[htbp]
  \centering
  \begin{overpic}[width=\linewidth, trim = 0 0 0 10, clip]{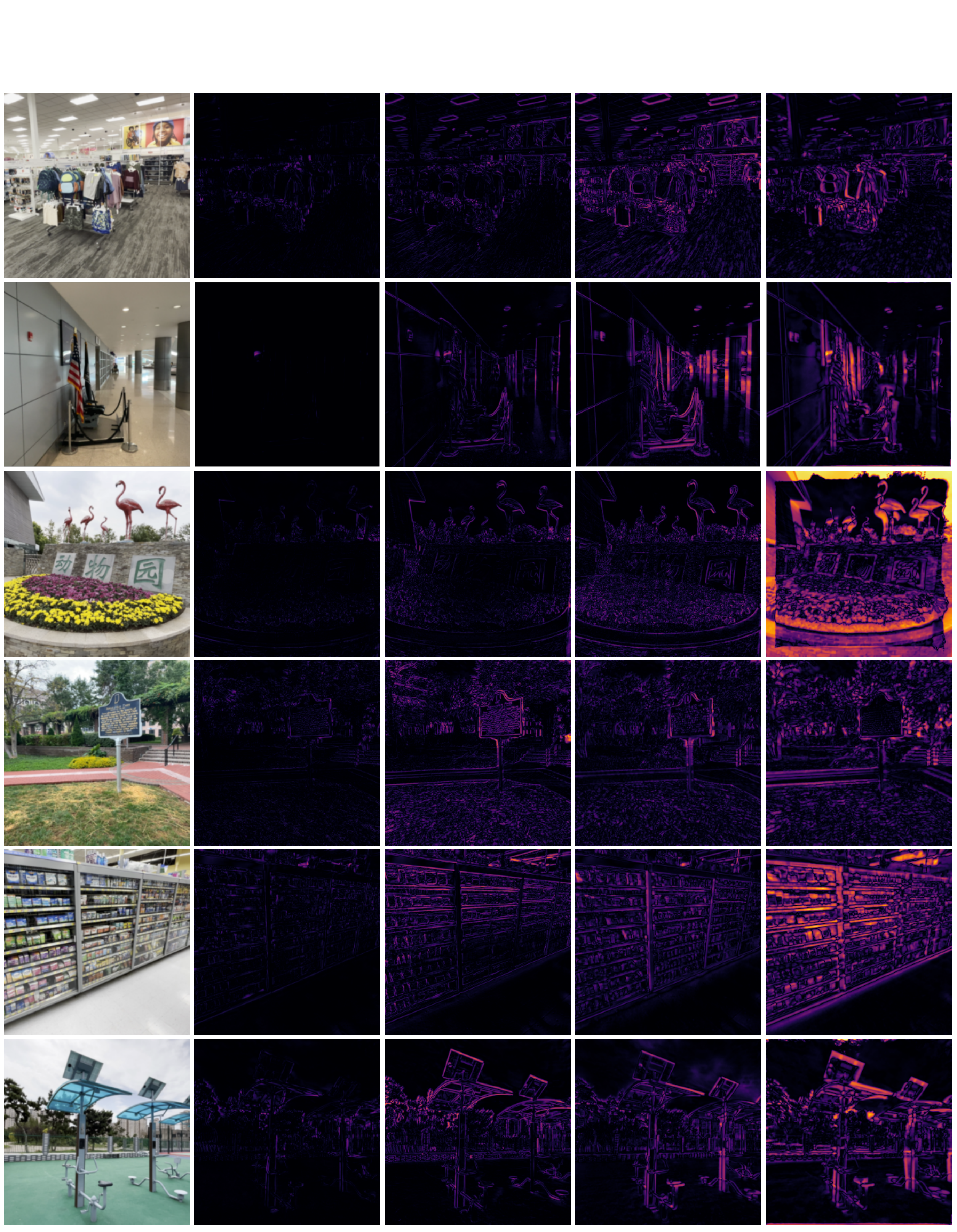}
    \put(6, 94){\textcolor{black}{GT}}
    \put(21.5, 94){\textcolor{black}{Ours}}
    \put(34, 94){\textcolor{black}{AnySplat~\cite{anysplat}}}
    \put(51, 94){\textcolor{black}{DA 3~\cite{da3}}}
    \put(65, 94){\textcolor{black}{Splatt3R~\cite{splatt3r}}}
  \end{overpic}
  \caption{\textbf{Qualitative comparison of L1 error maps.} We visualize the per-pixel L1 error between rendered images and the ground truth (GT). Darker regions indicate lower reconstruction errors. Compared to baseline methods (AnySplat~\cite{anysplat}, Depth Anything 3~\cite{da3}, and Splatt3R~\cite{splatt3r}), our StructSplat consistently exhibits significantly lower error across all scenes, demonstrating its superior capability in preserving texture details and maintaining high-fidelity reconstruction.}
  \label{fig:l1_error_map}
\end{figure}

\section{Additional Experiments}
\noindent\textbf{Larger-Scale Representation.} To evaluate the scalability of our approach, we further reduce the sampling rate of the source views, thereby covering a larger scene with fewer input views. The novel-view-synthesis results for this challenging setup are reported in Table~\ref{tab:larger_scene}. Across all three configurations (4, 6, 12, and 18 source views), our method consistently outperforms the baselines. Specifically, StructSplat achieves PSNR scores of 20.171~dB, 22.358~dB, 22.173~dB and 22.396~dB, with improvements of +3.319~dB, +4.233~dB, +4.286~dB and +3.476~dB over the second-best method, AnySplat~\cite{anysplat}. Furthermore, our approach achieves the best performance in terms of both SSIM and LPIPS. These results highlight StructSplat's strong robustness and its ability to reconstruct larger-scale scenes from sparse observations.

\noindent
\textbf{High Resolution Evaluation.} We conduct the quantitative comparison at the default resolutions of AnySplat~\cite{anysplat} (at 448$\times$448) and Depth Anything 3~\cite{da3} (at 504$\times$504). As shown in Tab.~\ref{tab:high_res}, StructSplat achieves the best performance across both high-resolution settings.

\noindent\textbf{Visualization of Rendered Depth.} In Figure~\ref{fig:depth}, we visualize the depth maps rendered from the 3D Gaussians predicted by our method. Notably, our model achieves precise depth estimation without relying on any ground-truth depth data or explicit depth consistency losses during training. Driven solely by RGB supervision, our proposed Structured Representation intrinsically captures the underlying 3D geometry, producing accurate and structurally consistent depth maps.

\noindent\textbf{Visualization of L1 Error Map.} Since visual differences between the rendered RGB images of different methods can sometimes be subtle, we additionally visualize the corresponding L1 error maps to explicitly highlight the performance gaps, as shown in Figure~\ref{fig:l1_error_map}. The error maps clearly indicate that the novel views synthesized by StructSplat exhibit substantially lower deviations from the ground truth. This reduction in error demonstrates our method's ability to recover fine texture details and preserve high-fidelity structural consistency across the scene.

\begin{figure}[t]
  \centering
  \includegraphics[width=0.95\linewidth]{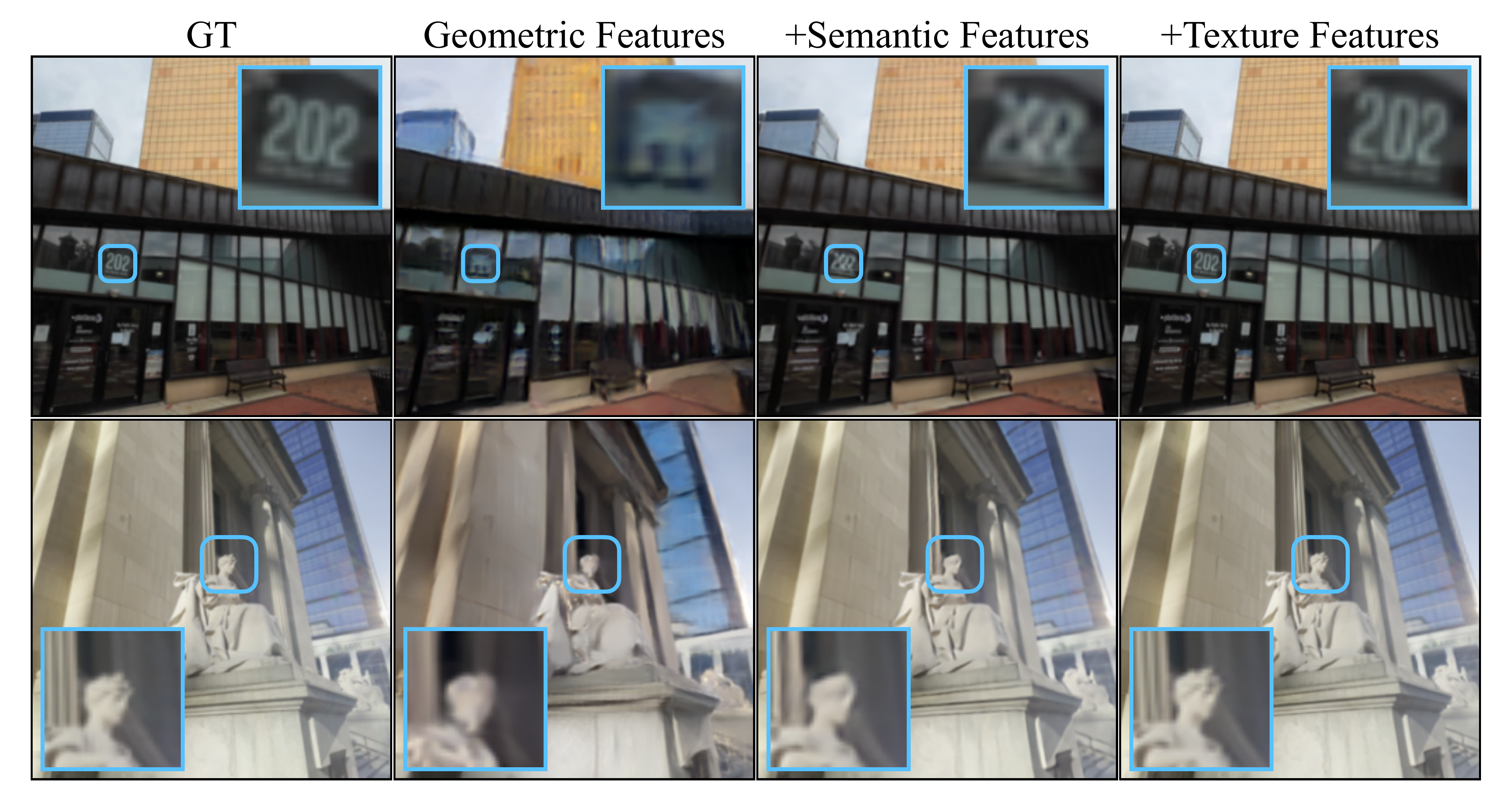}
  \caption{\textbf{Qualitative comparison of ablation.} 1) Using only geometric features causes severe structural collapse and large color artifacts. 2) Adding semantic features restores coherent structures and object layouts, while 3) further introducing texture features enables sharper images with accurate high-frequency details.
  \vspace{-0.3cm}
  }
  \label{fig:ab}
\end{figure}

\noindent
\textbf{Discussion on the Key Components. }
Quantitative results in Tab.~\ref{tab:ab_of_tf_sf} verify the effectiveness of our key components. We further conduct qualitative visual comparisons in Fig~\ref{fig:ab}. 1) Using only \textit{geometric features} causes severe structural collapse and large color artifacts. 2) Adding \textit{semantic features} restores coherent structures and object layouts, while 3) further introducing \textit{texture features} enables sharper images with accurate high-frequency details.

Regarding our choice and design of encoders: 1) we choose the VGGT \cite{vggt} backbone as the geometry encoder, and the Depth Anything 3 \cite{da3} backbone is also compatible with our framework. 2) VGGT \cite{vggt} fine-tunes DINOv2 for single-view geometric pre-extraction, resulting in a loss of appearance details in the geometry-oriented features. By introducing a separate frozen DINOv3 \cite{dinov3} encoder, we provide uncorrupted semantic priors. 3) We design a texture encoder to decouple Gaussian attributes (depth, opacity, color, scale, and rotation) to reduce interference among them.